\documentclass[a4paper,9pt,twocolumn,twoside]{article}

\usepackage[utf8]{inputenc}
\usepackage[scaled]{helvet}
\usepackage{mathptmx} 
\usepackage[english]{babel}
\usepackage[usenames,dvipsnames]{color}
\usepackage[hyperfootnotes=false]{hyperref}
\hypersetup{
 citecolor=red,
 linkcolor=blue,
 urlcolor=blue,
 colorlinks = true}

\usepackage{graphicx}
\usepackage{amsmath, amsfonts,amssymb}
\usepackage{array}

\usepackage{authblk}
\usepackage{sectsty}

\usepackage{geometry}
\usepackage[numbers,sort&compress]{natbib}
\sectionfont{\fontsize{13}{15}\selectfont}
\subsectionfont{\fontsize{11}{13}\selectfont}

\newcommand{\titlefont}{\normalfont\bfseries\fontsize{16}{19}\selectfont}

\allowdisplaybreaks[2]

\begin{document}

\title{\titlefont Efficient adaptation of complex-valued noiselet sensing matrices for compressed single-pixel imaging}

\author[1,*]{Anna Pastuszczak}
\author[2]{Bart\l{}omiej Szczygie\l{}}
\author[1]{Micha\l{} Miko\l{}ajczyk}
\author[1]{Rafa\l{} Koty\'{n}ski}

\affil[1]{University of Warsaw, Faculty of Physics, Pasteura 7, 02-093, Warsaw, Poland}
\affil[2]{University of Warsaw, College of Inter-Faculty Individual Studies in Mathematics and Natural Sciences, \.Zwirki i Wigury 93, 02-089 Warsaw, Poland}
\affil[*]{Corresponding author: apastuszczak@igf.fuw.edu.pl}

\date{\vspace{-3em}}

\twocolumn[
  \begin{@twocolumnfalse}
  \maketitle
  \renewcommand{\abstractname}{}
    \begin{abstract}
    \normalfont\fontsize{10}{13}\selectfont
     Minimal mutual coherence of discrete noiselets and Haar wavelets makes this pair of bases an essential choice for the measurement and compression matrices in compressed-sensing-based single-pixel detectors. In this paper we propose an efficient way of using complex-valued and non-binary noiselet functions for object sampling in single-pixel cameras with binary spatial light modulators and incoherent illumination. The proposed method allows to determine $m$ complex noiselet coefficients from $m+1$ binary sampling measurements. 
Further, we introduce a modification to the complex fast noiselet transform, which enables computationally-efficient real-time generation of the binary noiselet-based patterns using efficient integer calculations on bundled patterns.
The proposed method is verified experimentally with a single-pixel camera system using a binary spatial light modulator. 
\\ \\
    \end{abstract}
  \end{@twocolumnfalse}
]

\newcommand{\minus}{\scalebox{0.75}[1.0]{$-$}}


\section{Introduction}
Compressed sensing (CS)~\cite{CandesIntro, B_MathCs} is a technique of recovering a signal from an incomplete measurement based on an assumption that the signal has a sparse representation in a certain domain, for instance in some wavelet basis. In optics, CS has been initially applied for computational ghost imaging~\cite{Shapiro2008, Bromberg2009, Sun2012} giving rise to a novel image acquisition technique often referred to as single-pixel camera (SPC)~\cite{Baraniuk2008}, which allows for capturing images with a sole bucket detector rather than with a high-resolution array of detectors. This architecture opens the way for economic electro-optical imaging systems for infrared wavelengths~\cite{Gehm2015_EO_IR}, as well as for imaging in more exotic ranges of electromagnetic radiation, such as terahertz~\cite{Baraniuk_THz2008, Padilla_THz2014} or millimeter waves~\cite{Qiao2015_mm_hologr}.
Full color imaging~\cite{Dinozaury2013}, spectral imaging~\cite{Gehm2007_spectralimaging}, Stokes polarimetric imaging~\cite{OL_37_824_Duran}, and imaging of three-dimensional objects~\cite{Busck2004_laser3d,Li2015_laser3d,Javidi2012, Sun2013} have been also demonstrated. 
The range of research directions related to the use of CS expands rapidly, including lidar imaging~\cite{Lidar2012}, compressive holography~\cite{Lancis2013_hologr, Javidi2013_hologr}, sparse subwavelength imaging~\cite{Gazit2009_subwave, Eldar2011_subwave}, compressive pattern recognition~\cite{Nasze_statki_arcxiv}, rapid MRI diagnostics~\cite{CS_MRI_2008, Weizman2015_MRI} and imaging through scattering media such as a biological tissue~\cite{Lancis2015}.
Recently, a continuous real-time video recording at $10$~Hz with SPC has been also reported~\cite{Padget_sci_rep_2015}.


The common element of CS-based imaging techniques is the use of spatially modulated illumination or aperture. This modulation is frequently achieved with binary spatial light modulators (SLM), for instance with micromirror devices (DMD).
The choice of the sampling functions displayed by the SLM depends on the basis, in witch the sampled image has a sparse representation. Indeed, the minimal number of measurements necessary to recover the image is a function of the second power of the coherence between these two bases. 
As most of the real-world images are compressible in the wavelet domains, the application of sensing matrices incoherent with wavevelets into CS-based imaging systems is essential.
Presently, the sensing matrices are usually either based on Hadamard matrices or generated randomly. The Hadamard matrices are both easy to calculate and to display on a binary SLM, however their coherence with commonly used families of wavelets is rather high. The randomly generated patterns are acceptably incoherent with most of the image compression bases, however the necessity of storing the huge sensing matrix in the computer memory during the reconstruction of the image sets considerable limits on the resolution of the imaging system.

These problems are both overcome by sampling the images with noiselet functions~\cite{Coifman}.
Discrete noislets take values from four-element complex  sets, and importantly for CS with real-world images, are perfectly incoherent with Haar wavelet basis. A unitary noiselet matrix is fast to calculate. A way to calculate the fast noiselet transform (FNT) stems directly from the matrix definition.
These properties, which we will overview in more depth in Section~\ref{sec:noiselets}, decide upon the interest in the use of noiselets for SPC and are our motivation for the present work.

In this paper we focus on the efficient application of noiselet functions for SPC architectures.
In a SPC set-up with incoherent illumination, light carries a nonnegative real intensity signal. Together with a binary spatial modulation which is respectively represented with real binary functions, the use of more general complex CS measurement matrices is not straightforward. 
We propose an efficient method which allows to determine $m$ complex noiselet coefficients from $m+1$ measurements with non-negative binary sampling patterns.

\section{Compressed imaging}

\begin{figure}[t!]
\centering
	\includegraphics[width=8cm]{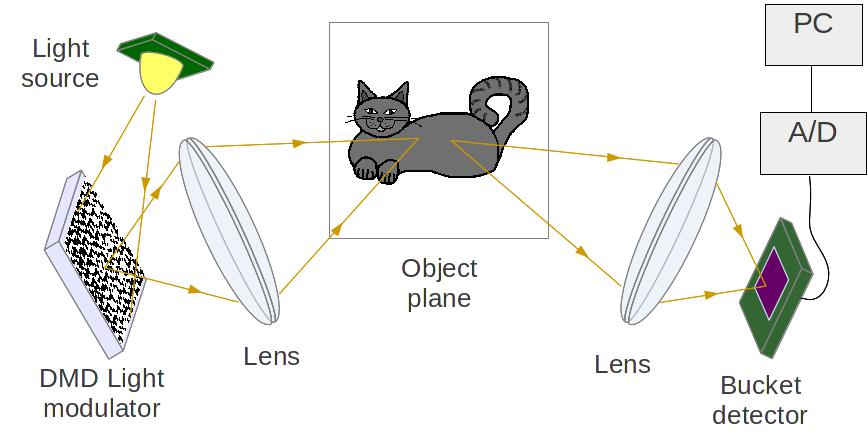}  
\caption{Schematic view of a single-pixel camera.}
\label{fig:schemat}
\end{figure}

A schematic view of a typical single-pixel sparse imaging set-up with incoherent illumination is illustrated in Fig.~\ref{fig:schemat}. The system consists of a light source, a SLM, imaging lenses, and a bucket detector. The SLM is used for illuminating the object plane with structured light consisting of a series of binary patterns. Simultaneously, the combined images of the patterns and the object are projected onto a single-pixel detector, which measures their total intensities. Following, the results are digitised, stored and further processed using a PC.

Let $Y$ be a vector of length $m$ containing all the measurements captured by the bucket detector. Then, every measurement $y_j$ ($1 \leq j \leq m$) within $Y$ is a dot product of a vector $\phi_j$ representing the brightness of the consecutive pixels of the $j$-th pattern displayed by the SLM and of a vector $X$ representing the reflectance (or transmittance) of the corresponding pixels of the image placed in the object plane:
\begin{equation}
y_j = \langle \phi_j, X \rangle.
\label{eq:single_measurement}
\end{equation}
A $m \times n$ matrix $\Phi$, whose rows consist of all the patterns $\phi_j$, is called a measurement or a sensing matrix.

In order to reconstruct the original image $X$ from the set of measurements $Y$, one needs to solve a system of linear equations: 
\begin{equation}
Y = \Phi\cdot X.
\label{eq:problem}
\end{equation}
In the case of under-sampled image sensing, i.e. when $m<n$, Eq.~(\ref{eq:problem}) has an infinite number of solutions. However, the reconstruction of the image is still possible, provided that the image has a sparse representation in a certain compression (sparsity) basis $\Psi$:
\begin{equation}
X = \Psi \cdot F,
\end{equation}
where $F$ is a vector of coefficients of $X$ in the basis $\Psi$. The vector $F$ can be either literally sparse, containing only a small number $S$ of non-zero elements, or all the elements apart from the $S$ largest ones can be negligible but still non-zero. In both cases the reconstruction of the image is performed by solving the basis pursuit optimisation problem~\cite{CandesIntro}:
\begin{equation}
\tilde F = \operatorname*{\arg\,min}_{F'} \Vert F' \Vert_1 \quad \textrm{subject to} \quad Y = \Phi{\cdot} \Psi{\cdot}  F'
\label{eq:min_l1}
\end{equation}
or, in case of noisy data acquisition, 
the basis pursuit denoise (BPDN) optimisation problem:
\begin{equation}
\tilde F = \operatorname*{\arg\,min}_{F'} \Vert F' \Vert_1 \quad \textrm{subject to} \ \ \Vert Y - \Phi {\cdot} \Psi {\cdot}  F' \Vert_2 < \varepsilon,
\label{eq:min_l1_noise}
\end{equation}
where $\Vert \cdot \Vert _{p}$ stands for the $\ell^p$ norm of a vector and $\varepsilon$ represents the level of noise present in the signal.
Alternative reconstruction approaches have been also successfully exploited, including minimisation of a certain $\ell^p$ quasi-norm for $0 \leq p <1$ \cite{Chartrand2007, Chartrand2008} or minimisation of the total variation (TV)~\cite{Chan2006_TV, Needell2013_TV, Ma2008_MRI_TV}.

The sufficient number of measurements required to collect enough data to reconstruct the original image without distortion satisfies the following inequality~\cite{CandesRomberg}:
\begin{equation}
m > C\cdot S\cdot  \log(n)\cdot \mu^2(\Phi, \Psi),
\label{eq:no_of_measurements}
\end{equation}
in which $C$ is a small constant, $S$ is the number of relevant coefficients in the compressed image $F$, and $\mu(\Phi, \Psi)$ is the mutual coherence of the sensing matrix $\Phi$ and the sparsity basis $\Psi$ defined as:
\begin{equation}
\mu(\Phi, \Psi) = \sqrt{n}\cdot \max_{j,k} \vert \langle \phi_j, \psi_k \rangle \vert,\label{eq:coherence} 
\end{equation}
where $\phi_j, \psi_k$ (for $1 \leq j \leq m$ and $1 \leq k \leq n$) stand for the row vectors of the matrices $\Phi$ and $\Psi$ respectively. 
Therefore, to recover the original image from the least possible number of measurements, it is crucial to choose the sensing matrix $\Phi$ in such a manner, that the mutual coherence $\mu(\Phi, \Psi)$ is kept as small as possible.
In other words, the sensing matrix should be almost completely incompressible in the basis $\Psi$.


\section{Noiselet matrices}
\label{sec:noiselets}

In 2001 a family of functions was introduced, named noiselet functions~\cite{Coifman}, which is perfectly incoherent with the Haar wavelet basis (mutual coherence between noiselet and Haar orthonormal basis equals $1$). Since most of the real-life images are well compressible in the Haar wavelet domain, noiselets are then a good candidate for constructing an efficient sensing matrix.

Another advantage of the discrete noiselet-based sensing matrices over e.g. the Gaussian random matrices is that they are defined using a recursive formula based on the Kronecker product $\otimes$ (a similar formula was introduced in~\cite{Tuma_Hurley_noiselets}): 
\begin{equation}
\begin{aligned}
N_1 &= \begin{bmatrix} 1 \end{bmatrix}, \\
N_{2n} &= \frac{1}{{2}} \begin{bmatrix} 1-i & 1+i \\ 1+i & 1-i \end{bmatrix} \otimes N_{n}.
\end{aligned}
\label{eq:noiselet_matrix}
\end{equation}
where $N_n$ are $n \times n$ unitary  matrices whose dimension is a power of two $n=2^q$ (for $q=0,1,2,...$). It is worth mentioning, that Eq.~(\ref{eq:noiselet_matrix}) defines the matrices of both one-dimensional and two-dimensional noiselet transforms. 
Indeed, the 2D noiselet transform of a $n \times k$ matrix $A$ takes the form:
\begin{multline}
[N^{2D}_{n \times k}] \cdot \textrm{vec}(A) = \textrm{vec}(N_n \cdot A \cdot N_k^T) = \\ = (N_k \otimes N_n) \cdot \textrm{vec}(A)  =  [N_{n \times k}] \cdot \textrm{vec}(A),
\label{eq:2D_noiselets}
\end{multline}
where $\textrm{vec}(A)$ denotes vectorisation of matrix $A$ obtained by stacking all columns of $A$ into a single column vector.
The second equality in Eq.~(\ref{eq:2D_noiselets}) is a well known property of the Kronecker product: $\textrm{vec}(ABC^T) = (C \otimes A) \textrm{vec}(B)$ for matrices $A,B,C$, whereas
the last equality is a straightforward consequence of the associativity of the Kronecker product, which is recursively used to construct the noiselet matrices (see Eq.~(\ref{eq:noiselet_matrix})).

 Therefore, the operation of multiplication by a noiselet matrix is replaceable with a fast transform with a computational complexity of $\mathcal{O}(n \log n)$ for both one-dimensional and two-dimensional transforms. Additionally, the huge $n\times n$ noiselet matrix is actually never constructed during the evaluation of the matrix-vector product and hence does not need to be stored in computer memory.  

Ignoring the normalizing factors, the elements of the noiselet matrices take discrete values from one of two 4-element sets, depending on the parity of the parameter $q$ defining the size of the matrix $N_n=N_{(2^q)}$:
\begin{equation}
\begin{aligned}
\sqrt{n} \cdot N_n(j,k)   & \in \lbrace  1, \ \minus 1, \ i, \ \minus i \rbrace                   &\textrm{ for even } q  \\
\sqrt{2n} \cdot N_n(j,k) & \in \lbrace  1{+}i, \ 1{-}i, \ \minus 1{+}i, \ \minus 1{-}i \rbrace    &\textrm{ for odd } q ,  
\end{aligned}
\label{eq:elem_sets}
\end{equation}
where $1 \leq j,k \leq n$ are indices of an element of the matrix $N_{n}$.


Owing to these properties, noiselet matrices have an excellent potential of being used as the sensing matrices in CS imaging systems. However, displaying complex-valued patterns with the use of the DMD is impossible, or at least not straightforward.
A solution to this problem was proposed recently~\cite{zhao2015}, suggesting to divide the complex-valued sensing matrix into four separate matrices, namely: 1. the positive real part, 2. the negative real part, 3. the positive imaginary part, and 4. the negative imaginary part, and to perform the measurements with each of these matrices independently. Then, the collected data may be synthesized into an equivalent of a single sequence of measurements obtained with the complex-valued sensing matrix.
This method however, suffers from 
that the number of snapshots taken by the SPC is fourfold larger than the number of samples actually measured, unnecessarily prolonging the data acquisition time. Instead, we want to replace the complex noiselet functions with the same number of real binary functions, but to retain the incoherence of the basis with Haar wavelets.

 In conclusion we note that the definition of discrete noiselet transform is very similar to that of Walsh-Hadamard transform, which also shares the same form for one-dimensional and two-dimensional case, has a similar fast calculation method, and is also commonly used in CS applications especially that it is readily binary. However, the mutual coherence of Walsh-Hadamard matrices with Haar wavelets is much larger.

\section{Efficient method of displaying noiselet-based sensing matrices}
\label{sec:patterns}
We propose an efficient method of displaying noiselet-based sensing matrices with the use of a DMD for the purpose of compressive imaging. The method allows for obtaining $m$ complex-valued measurement samples by modulating the object with exactly $m+1$ real binary patterns.
We note, that in general the DMDs enable displaying also grayscale patterns with multiple intensity levels by flickering of the micromirrors, however the use of binary patterns is the most efficient in terms of the frequency of pattern exposure and the stability of the displayed images.

The proposed procedure takes the following general form:
\begin{enumerate}
\item The sensing matrix $\Phi$ is composed of $m$ rows chosen randomly from a complex-valued noiselet matrix $N_n$. \label{proc:1}
\item A real binary matrix $P$ (further in this paper called a pattern matrix) is defined, whose rows are linear functions of the rows of matrix $\Phi$.
\item A series of measurements $\tilde{Y}$ is taken by the SPC using the row vectors $p_j$ of the matrix $P$ as the patterns displayed by the DMD:
\begin{equation}
\tilde{Y} = P\cdot X.
\end{equation}
\item The complex-valued vector $Y$ is calculated from the measurements $\tilde{Y}$ and used for reconstructing the image $X$.
\end{enumerate}

\begin{table}[tb!]
\centering
\caption{\bf Examples of noiselet matrices $N_{2^q}$ with either even or odd value of parameter $q$.}
\begin{tabular}{ m{1.4cm}  c  c  }
   \hline  & $4 \cdot N_{16} \quad (q=4)$ & $8 \cdot N_{32} \quad (q=5)$ \\ \hline
  \begin{center} Real part  \end{center} 
                                        & \raisebox{-0.5\height}{\includegraphics[width=2.9cm]{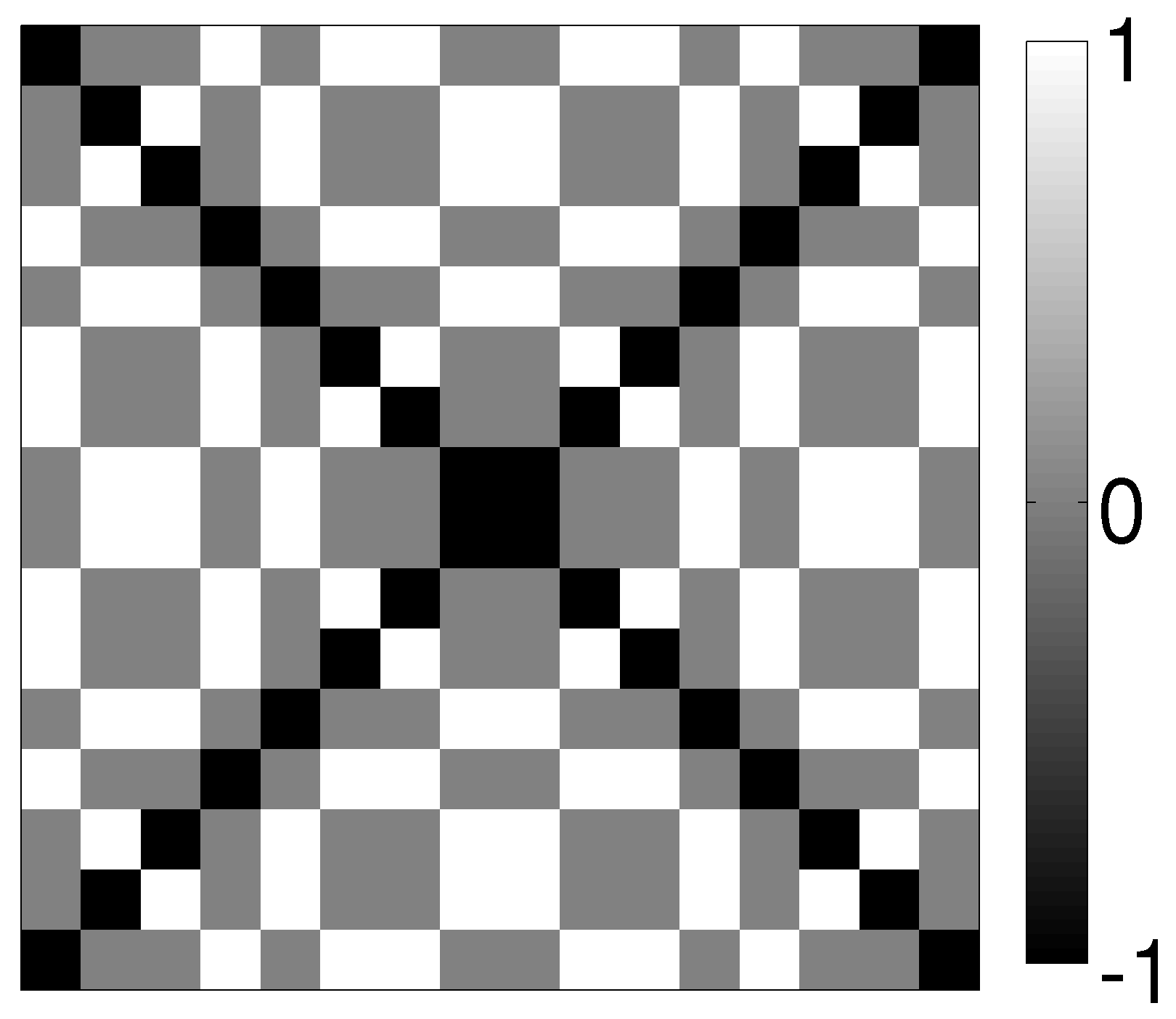}} & 
  										 \raisebox{-0.5\height}{\includegraphics[width=2.9cm]{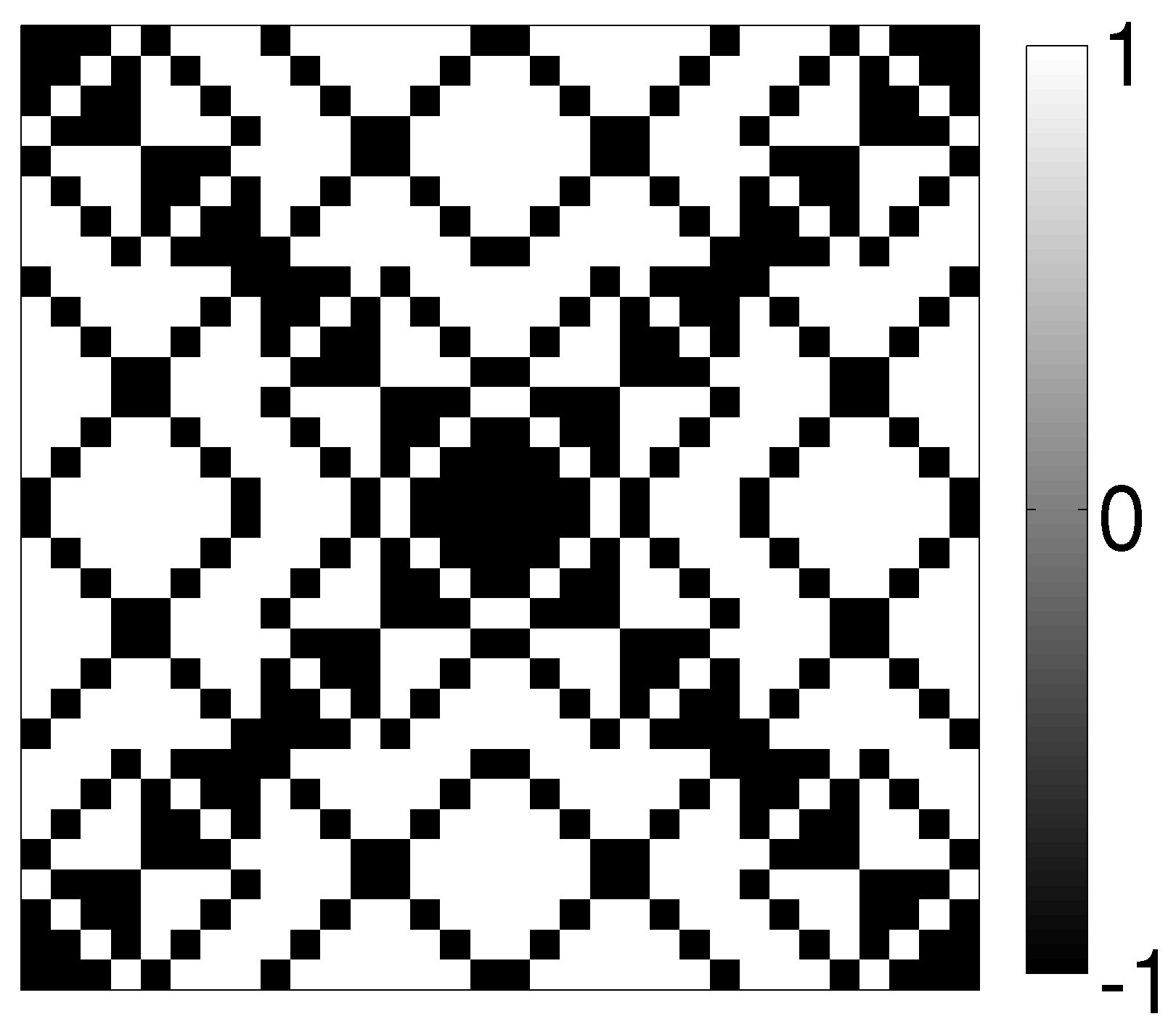}} \\ \hline
  \begin{center} Imaginary part \end{center}
                                        & \raisebox{-0.5\height}{\includegraphics[width=2.9cm]{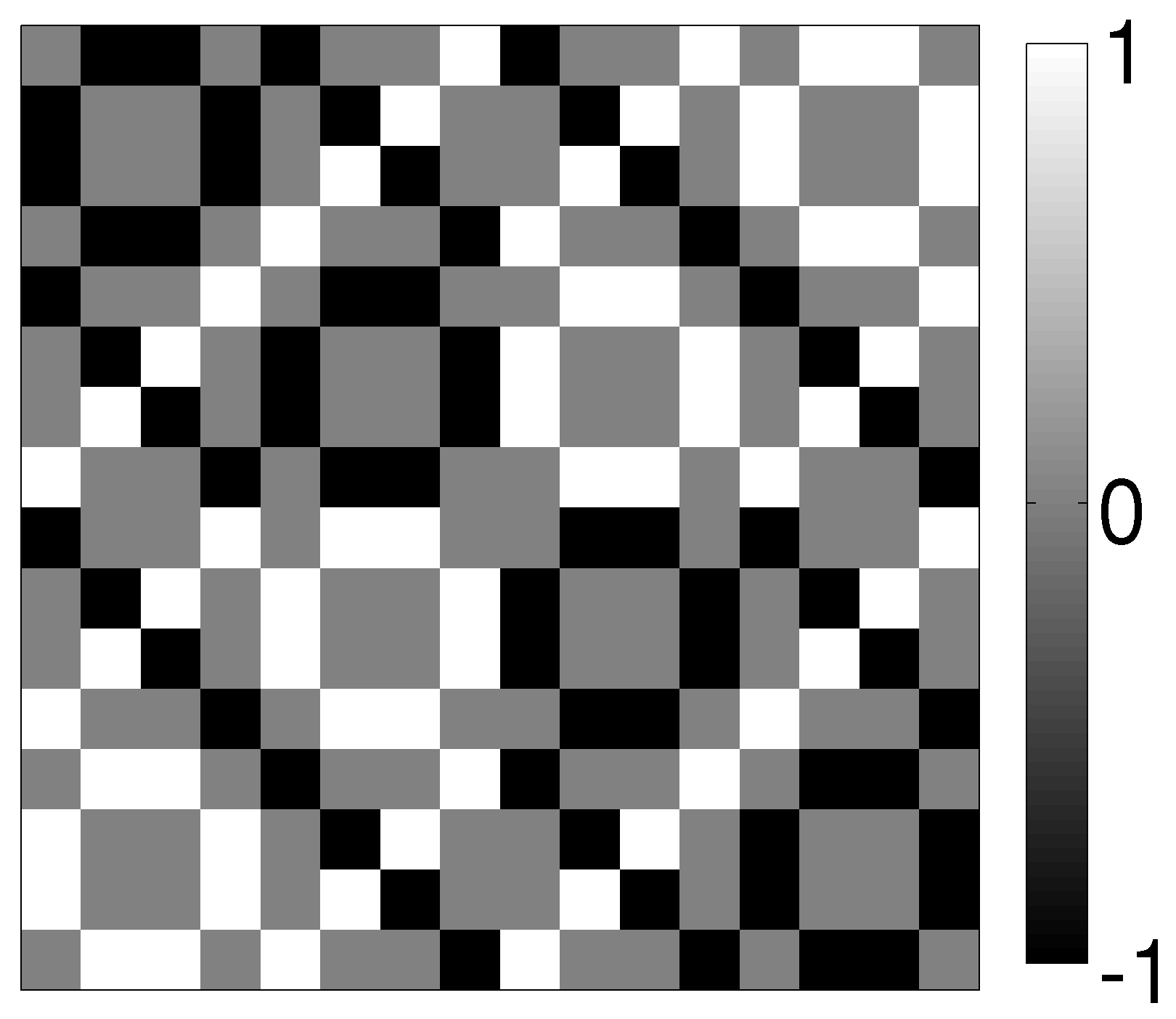}} & 
  									     \raisebox{-0.5\height}{\includegraphics[width=2.9cm]{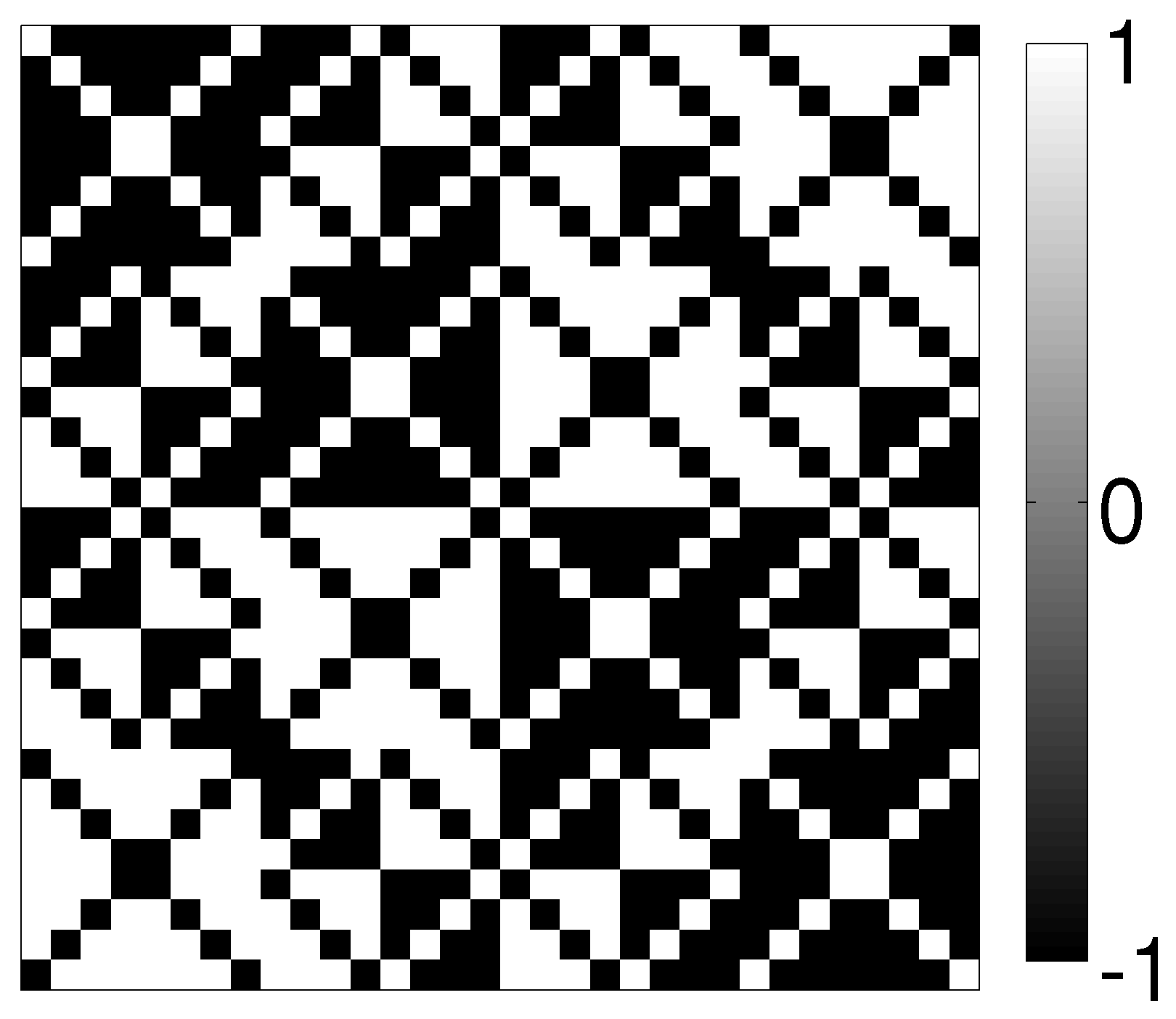}} \\ \hline
  \begin{center} Sum of real and imaginary parts \end{center} 
                                        & \raisebox{-0.5\height}{\includegraphics[width=2.9cm]{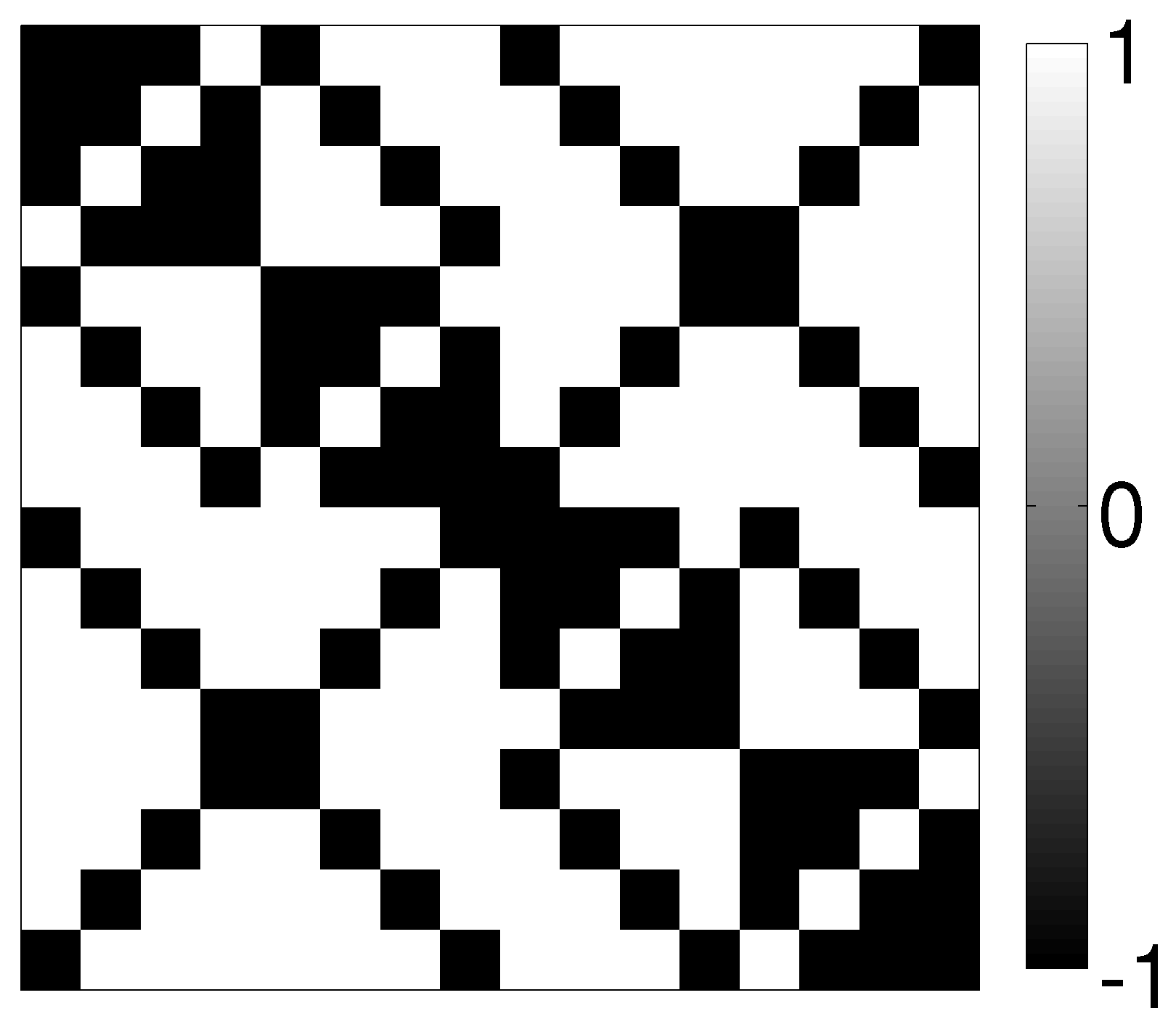}}  & 
                                          \raisebox{-0.5\height}{\includegraphics[width=2.9cm]{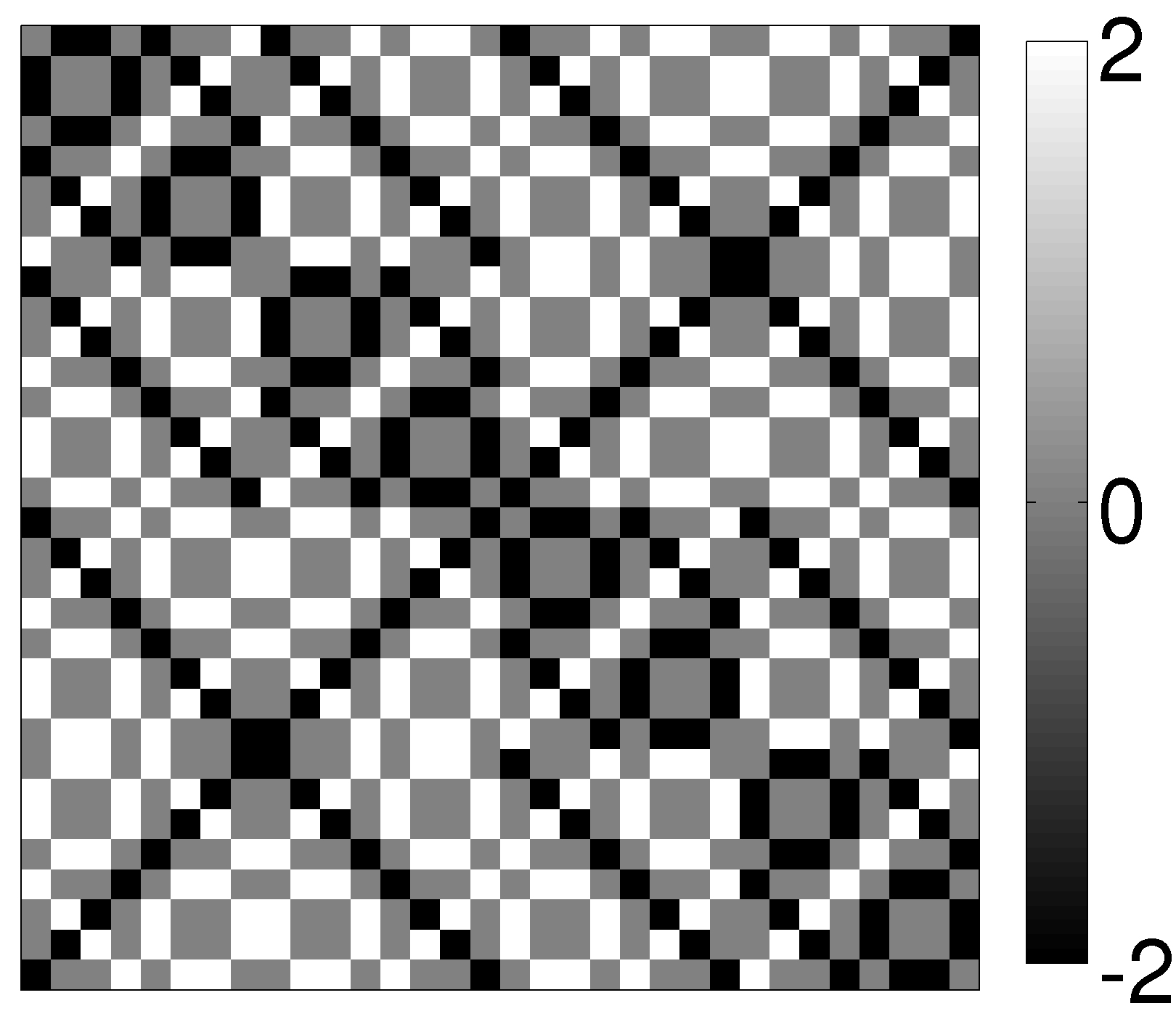}}  \\ \hline
  \begin{center} Difference of real and imaginary parts \end{center} 
                                        & \raisebox{-0.5\height}{\includegraphics[width=2.9cm]{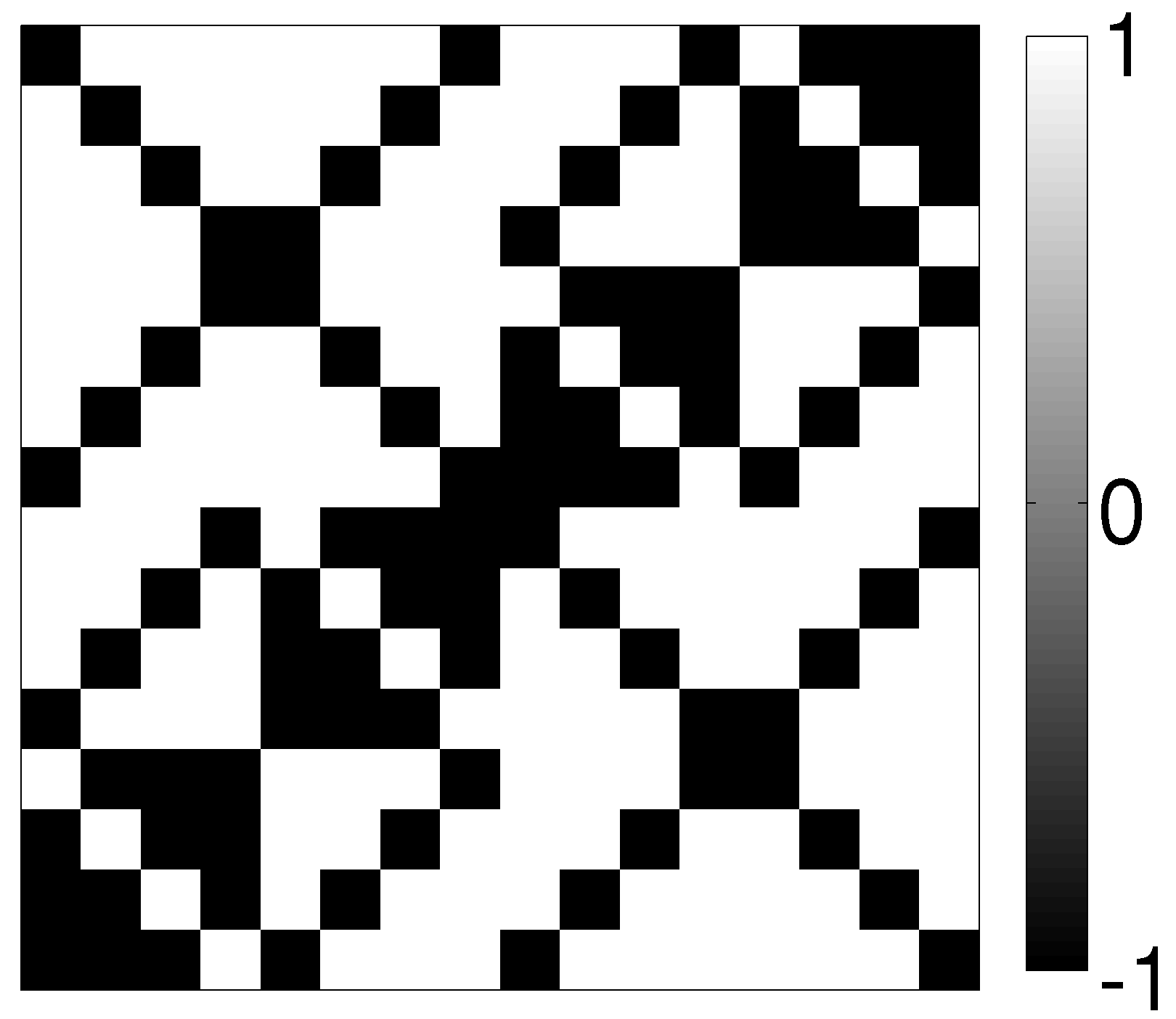}} & 
                                          \raisebox{-0.5\height}{\includegraphics[width=2.9cm]{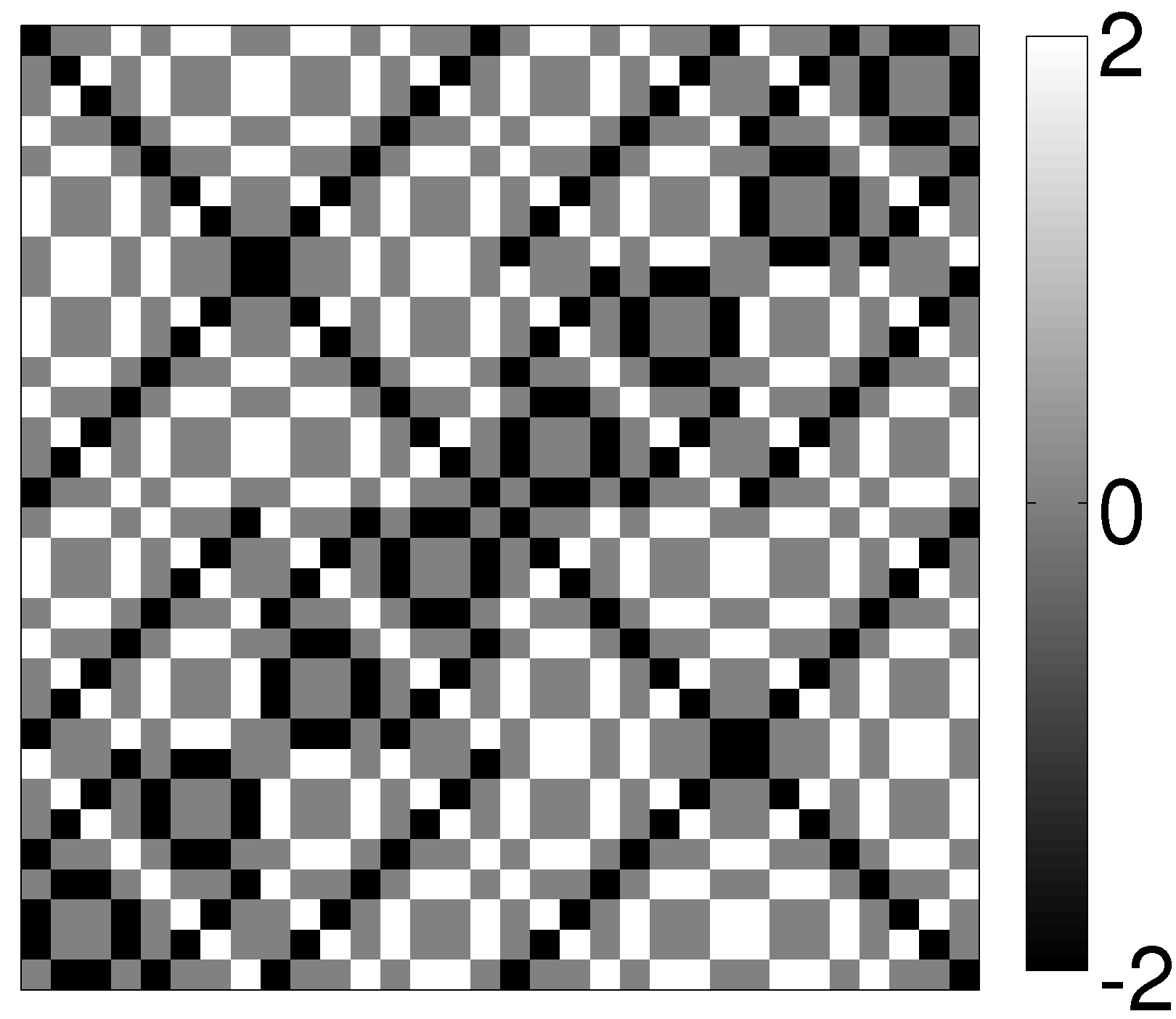}} \\ \hline
\end{tabular}
\label{tab:noiselets}
\end{table}

In order to establish the form of the matrix $P$ we exploit several properties of the noiselet matrices. 
First we note, that for odd values of parameter $q$, real and binary patterns are obtained immediately from the real and imaginary part of the noiselet matrix $N_{2^q}$ (see Eq.~\ref{eq:elem_sets}). For even values of the parameter $q$, the real and the imaginary part of a noiselet matrix are triple-valued, however binary patterns are obtained from their sum and difference instead. For better illustration, two examples of a noiselet matrix with either even or odd value of the parameter $q$ are presented in Table~\ref{tab:noiselets}.

However, thus obtained set of patterns suffers from two important shortcomings: 1. the number of real-valued patterns to be displayed by the DMD is twice as large as the number of complex-valued patterns, which they represent and 2. the patterns consist of both positive and negative values. We shall address these issues consecutively.

\begin{figure}[tb!]
\centering
\includegraphics[width=8.5cm]{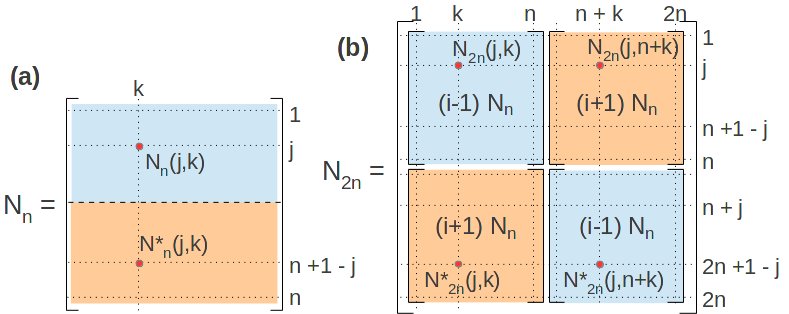}         
\caption{Reflection symmetry of the noiselet matrix. Any element taken from the upper half of the matrix  is a complex conjugate of the mirror element taken from the lower half of the matrix. Matrix (b) is obtained from matrix (a) using formula introduced in Eq.~\ref{eq:noiselet_matrix}.}
\label{fig:macierz}
\end{figure}

We begin with the first. 
This problem may be solved by exploiting the symmetry of the noiselet matrices. 
Noiselet matrices, similarly to discrete Fourier transform matrices, obey the reflection symmetry of the form:
\begin{equation}
\forall_{1 \leq j,k \leq n} \quad N_n^\ast(j,k) = N_n(n{+}1{-}j, k),
\label{eq:mirror-symmetry}
\end{equation}
where the symbol $^\ast$ denotes the complex conjugate. In other words, each element taken from the upper half of a noiselet matrix $N_n$ (see Fig.~\ref{fig:macierz}(a)) is a complex conjugate of the respective mirror element taken from the lower half of the matrix. 

Indeed, suppose that Eq.~(\ref{eq:mirror-symmetry}) is fulfilled for a noiselet matrix $N_n$. Then, by induction, for the matrix $N_{2n}$ (see Fig.~\ref{fig:macierz}(b)), the elements taken from the left half of the matrix satisfy :
\begin{multline}
\begin{aligned}
N_{2n}^\ast(j,k) &= [(1{-}i) N_n(j,k)]^\ast = (1{+}i) N_n^\ast(j,k) = \\&= (1{+}i)N_n(n{+}1{-}j, k) = N_{2n}(2n{+}1{-}j, k),
\end{aligned}
\end{multline}
where the first and the last equality result from the recursive definition of the matrix $N_{2n}$ (Eq.~(\ref{eq:noiselet_matrix})) and the third equality results from Eq.~(\ref{eq:mirror-symmetry}).
\\
Similarly, for the elements from the right half of the matrix $N_{2n}$:
\begin{multline}
\begin{aligned}
N_{2n}^\ast(j,n{+}k) = [(1{+}i) N_n(j,k)]^\ast = (1{-}i) N_n^\ast(j,k) = \\= (1{-}i)N_n(n{+}1{-}j, k) = N_{2n}(2n{+}1{-}j, n{+}k).
\end{aligned}
\end{multline}

Owing to this property, a pair of patterns consisting of the real and of the imaginary part of a single noiselet (i.e. of a single row vector taken from a noiselet matrix) contains full information about two mirror noiselets taken from the upper and the lower half of the noiselet matrix respectively. 
Therefore, we propose to choose the sensing matrix $\Phi$ in such a manner that it consists of $m/2$ rows randomly picked from the upper half of a noiselet matrix $N_n$ and the $m/2$ mirror rows taken from the lower half of the matrix. Then, the number of unique pattens composed of the real and the imaginary parts (or of their sum and difference) of the rows of matrix $\Phi$ equals $m$.

The second problem, concerning patterns consisting of both positive and negative values, is resolved by applying additional rescaling to the patterns. The patterns are already binary, therefore the necessity of displaing negative-valued pixels may be omitted by replacing them with zeros instead. Thus obtained patterns consist of values $0$ and $1$ only and they are straightforward to be displayed by a DMD. 
The cost of this rescaling is the necessity of capturing a single additional measurement, independently of the number of patterns $m$ or of the size of  image $n$. To justify this statement let us first analyse the procedure of restoring the complex-valued measurement vector $Y$ from the real-valued measurements $\tilde{Y}$.

Let us order  the row-vectors $\phi_j$ of the sensing matrix $\Phi$ to obey the following formula:
\begin{equation}
\phi_j = \phi^\ast_{m+1-j}, \quad \textrm{where} \ j=1,2,3,...,m/2.
\end{equation} 

The patterns, i.e. the rows of the proposed pattern matrix $P$, take one of two forms, depending on the parity of the parameter $q$ defining the size of the sampled image~($n=2^q$): 
\begin{equation}
\begin{aligned}
\text{for odd }     &q : \\
p_{2j-1}            &= \tfrac{1}{2} \big(\sqrt{2n}\ \Re(\phi_j) + 1 \big), \\
p_{2j}              &= \tfrac{1}{2} \big(\sqrt{2n}\ \Im(\phi_j) + 1 \big), \\[5pt]
\text{for even }     &q : \\
p_{2j-1}             &= \tfrac{1}{2} \big(\sqrt{n}\ [\Re(\phi_j) + \Im(\phi_j)] + 1 \big), \\
p_{2j}               &= \tfrac{1}{2} \big(\sqrt{n}\ [\Re(\phi_j) - \Im(\phi_j)] + 1 \big), \\[5pt]
\text{where} \quad j &= 1,2,3,...,m/2.
\end{aligned}
\label{eq:patterns_def}
\end{equation}
In both cases, all the patterns stored in the matrix $P$ are real and consist of zeros and ones only.

Simultaneously, the row vectors of the sensing matrix $\Phi$ are expressed in terms of the patterns $p_{2j-1}, p_{2j}$ by the following equations:
\begin{equation}
\begin{aligned}
\text{for odd }      &q : \\
\quad \phi_j         &=  \tfrac{1}{\sqrt{2n}}\ \big(2p_{2j-1} +2i p_{2j} - (1{+}i) \big), \\[5pt]
\text{for even }     &q : \\
\quad \phi_j         &=  \tfrac{1}{\sqrt{n}}\  \big((1{+}i)p_{2j-1} + (1{-}i) p_{2j} - 1 \big), \\[5pt]
\text{where} \quad j &= 1,2,3,...,m/2.
\end{aligned}
\label{eq:patterns_reverse}
\end{equation}

Finally, the equivalents of the complex-valued measurements~$y_j$ are restored from the actual real-valued measurements~$\tilde y_k$ using the following formulas: 
\begin{align*}
\text{for}& \text{ odd }  q : \\
y_j        &= \ y^\ast_{m+1-j} = \langle \phi_j, X \rangle  
=\\        &= \tfrac{1}{\sqrt{2n}}\   \Big(2\langle p_{2j-1},X \rangle  + 2i\langle p_{2j},X \rangle  - (1{+}i)\langle I_v,X \rangle \Big)
=\\        &= \tfrac{1}{\sqrt{2n}}\   \Big(2 \ \tilde{y}_{2j-1} + 2i \ \tilde{y}_{2j} -(1{+}i)\langle I_v,X \rangle \Big), 
\displaybreak \\[5pt]
\text{for}& \text{ even }  q : \\
y_j        &= \ y^\ast_{m+1-j} = \langle \phi_j, X \rangle 
=\\        &=  \tfrac{1}{\sqrt{n}}\    \Big((1{+}i)\langle p_{2j-1},X \rangle + (1{-}i)\langle p_{2j},X \rangle -\langle I_v,X \rangle \Big)
=\\        &= \tfrac{1}{\sqrt{n}}\    \Big((1{+}i) \ \tilde{y}_{2j-1} + (1{-}i) \ \tilde{y}_{2j} -\langle I_v,X \rangle \Big), \\[5pt]
           &\text{where} \quad j = 1,2,3,...,m/2,
\stepcounter{equation}\tag{\theequation}\label{eq:Y_recovery} 
\end{align*}

and $I_v$ denotes a vector of length $n$ with all elements equal to $1$. 
The term $\langle I_v,X \rangle$, which occurs in both cases in Eq.~(\ref{eq:Y_recovery}), is the consequence of the rescaling applied to the patterns in order to obtain only non-negative values. In physical interpretation, this term represents a measurement of the total intensity of the image $X$ without modulating it with any pattern. This single additional measurement is necessary for reverting the scaling during retrieval of $Y$. Therefore, exactly $m+1$ measurements with real binary and non-negative patterns are required in order to restore $m$ samples corresponding to measurements taken with a complex-valued and non-binary noiselet-based sensing matrix. 

In practical experimental conditions this number of measurements may be increased. For instance, in order to eliminate the background `dark pixel' signal resulting from the light reflected from the DMD matrix when all the mirrors are in the off-state, it is necessary to measure that dark signal in the calibration stage, and then to subtract it from the measurements. More likely, the intensity of the light-source may vary with time. An additional detector could be used for the normalization of measurements \cite{Sun2012}. In another approach, a differential measurement with complementary~\cite{Yu2015_complementary3D} binary masks $(\langle p_k,X \rangle-\langle I_v{-}p_k,X \rangle)/2$
 increases the signal-to-noise-ratio, eliminates background dark signal, and accounts for intensity variations but actually doubles the number of measurements.


\section{Modification of the fast noiselet transform for real-time generation of the noiselet-based patterns}
\label{sec:patterns_real_time_gen}
In the following section we propose a modification of the fast noiselet transform for the purpose of efficient generation of the noiselet-based real binary patterns introduced in Section~\ref{sec:patterns}. 
The one-dimensional and two-dimensional noiselet transform defined by Eq.~(\ref{eq:noiselet_matrix}) operates on complex numbers. We propose a similar procedure, allowing for a direct generation of the real binary patterns $p_j$ using only operations of summation and subtraction on integer variables.
Moreover, by utilising the bit representation of a $k$-bit integer, a bundle of up to $k{-}2$ patterns $p_j$ may be generated using the same number of arithmetic operations as in the case of generating only a single pattern (the additional two bits are used in the intermediate calculations).
The efficiency of the proposed method allows for generating the patterns $p_j$ during the time of the experiment, without any preparations beforehand.

The modified noiselet transform matrix $\tilde N_n$ (where $n = 2^q$) relates to the noiselet matrix $N_n$ as follows:
\begin{equation}
\tilde N_n = \sqrt{2n} \cdot N_n \exp(i \tfrac{\pi}{4}(q+1))
\label{eq:noiselet_matrix_modified_relative}
\end{equation}
and it satisfies a recursive formula similar to that, which defines matrices $N_n$:
\begin{equation}
\begin{aligned}
\tilde N_1 &= \begin{bmatrix} 1+i \end{bmatrix}, \\
\tilde N_{2n} &= \begin{bmatrix} 1 & i \\ i & 1 \end{bmatrix} \otimes \tilde N_{n}.
\end{aligned}
\label{eq:noiselet_matrix_modified}
\end{equation}
We note, that the elements of thus defined matrix $\tilde N_n$ belong to a single 4-element set, independently on the parameter $q$:
\begin{equation}
\tilde N_n(j,k)  \in \lbrace  1{+}i, \ 1{-}i, \ \minus 1{+}i, \ \minus 1{-}i \rbrace.
\label{eq:elem_set_modified}
\end{equation}

To derive the modified noiselet transform $\mathcal{\tilde N}$, let us rewrite Eq~(\ref{eq:noiselet_matrix_modified}) into a more explicit form. We define an auxiliary matrix:
\begin{equation}
\tilde N_G = \begin{bmatrix} 1 & i \\ i & 1 \end{bmatrix}.
\label{eq:noiselet_matrix_generator}
\end{equation}
Then, the modified noiselet matrix takes a form:
\begin{equation}
\begin{aligned}
&\tilde N_{2^q} = (1{+}i) \underbrace{ \tilde N_G \otimes \tilde N_G \otimes ... \otimes \tilde N_G }_q =\\
               &= (1{+}i) (\tilde N_G \otimes I_2 \otimes ... \otimes I_2) \cdot \\
               &\quad \cdot (I_2 \otimes \tilde N_G \otimes I_2 \otimes ... \otimes I_2) ...
                        (I_2 \otimes ... \otimes I_2 \otimes \tilde N_G) = \\
               &= (1{+}i) (\tilde N_G \otimes I_{2^{q-1}}) (I_2 \otimes \tilde N_G \otimes I_{2^{q-2}}) ... 
                        (I_{2^{q-1}} \otimes \tilde N_G),
\end{aligned}
\label{eq:noiseles_matrix_modified_explicit}
\end{equation}
where $I_n$ stands for an identity matrix of size $n \times n$, and the second equality results from the mixed-product property of the Kronecker product: $(A \otimes B)(C \otimes D) = (AC)\otimes (BD)$.
Each factor in the final expression in Eq.~(\ref{eq:noiseles_matrix_modified_explicit}) of a form $I_{2^{k-1}} \otimes \tilde N_G \otimes I_{2^{q-k}}$ for $k = 1,2,..., q$, is a block diagonal matrix with $2^{k-1}$ same blocks of the form:
\begin{equation}
\tilde N_G \otimes I_{2^{q-k}} = \begin{bmatrix}  I_{2^{q-k}} & i\ I_{2^{q-k}} \\ i\ I_{2^{q-k}} & I_{2^{q-k}} \end{bmatrix}.
\label{eq:noiselet_matrix_block}
\end{equation}
Therefore, the modified noiselet transform of a vector (i.e. the multiplication of the vector by the matrix $\tilde N_{2^q}$) is equivalent to dividing the vector into $2^{k-1}$ blocks, each of length $2^{q-k+1}$, and multiplying each of them by the matrix (\ref{eq:noiselet_matrix_block}). The procedure is then repeated $q$ times for sizes of the blocks defined by consecutive values of $k$. The product of a single partial vector $v = [\begin{smallmatrix} u\\ w \end{smallmatrix}]$ by the matrix (\ref{eq:noiselet_matrix_block}) has a simple form:
\begin{equation}
\begin{bmatrix}  I_{2^{q-k}} & i\ I_{2^{q-k}} \\ i\ I_{2^{q-k}} & I_{2^{q-k}} \end{bmatrix} \ 
\begin{bmatrix} u \\ w \end{bmatrix} = \begin{bmatrix} u + i w \\ w + i u \end{bmatrix},
\end{equation}
which, by keeping the real and the imaginary parts of the vectors as separate variables, does not require any implementation of complex numbers. 
Moreover, the only arithmetical operations present in the transform are summations and subtractions applied to the real and the imaginary part of the vector $v$.
Therefore, for vector $v$ consisting only of integer values, the transform may be implemented purely on integer types, which greatly increases its efficiency.

The real binary patterns $P$ introduced in the previous section 
are obtained directly from the modified noiselet transform. 
Indeed, if a complex sensing pattern $\phi_j$ is chosen as the $k$-th row of the noiselet matrix $N_n$, then the pair of its equivalent real binary patterns $p_{2j-1}$, $p_{2j}$ defined by Eq.~(\ref{eq:noiselet_matrix_modified_relative}) is expressed by the modified noiselets as follows:
\begin{equation}
\begin{aligned}
\text{for } q \bmod 4 \ \geq 2  \\     
p_{2j-1}     &= \pm \ a_k,     \\
p_{2j}       &= \pm \ b_k,    \\
\text{for } q \bmod 4 \  < 2   \\        
p_{2j-1}     &= \pm \ b_k,    \\
p_{2j}       &= \pm \ a_k,     \quad j=1,2,3,...,m/2,
\end{aligned}
\label{eq:patterns_mod_noiselets_relation}
\end{equation}
where
\begin{equation}
\begin{aligned}
a_k &= \tfrac{1}{2}(\Re(\tilde N_{n, k}) +1), \\
b_k &= \tfrac{1}{2}(\Im(\tilde N_{n, k}) +1),
\end{aligned}
\end{equation}
$\tilde N_{n, k}$ indicates the $k$-th row of the matrix $\tilde N_n$ and $ q \bmod a$ stands for the reminder after division of parameter $q$ by~$a$.
The signs in the expressions for $p_{2j-1}$, $p_{2j}$ depend on the value of the complex argument 
in Eq.~(\ref{eq:noiselet_matrix_modified_relative}).

Now, let us explain the operation of the packed modified noiselet transform. A single pattern $a_k$ or $b_k$ is efficiently obtained by applying the modified noiselet transform to a unit vector $e_k$, whose all elements apart from the $k$-th one are zeros:
\begin{equation}
\begin{aligned}
a_k = \tfrac{1}{2} \Re(\mathcal{\tilde N} (e_k) +1), \\
b_k = \tfrac{1}{2} \Im(\mathcal{\tilde N} (e_k) +1).
\end{aligned}
\label{eq:transf_ak_bk}
\end{equation}
The operation required to obtain a bundle of different binary patterns $a_{k_j}$ (or $b_{k_j}$ with interchanging $\Re\leftrightarrow \Im$) encoded into the consecutive bit-planes of the bit representation of the integer variables is derived from Eq.~(\ref{eq:transf_ak_bk}):
\begin{equation}
\begin{aligned}
a_{\text{packed}} 
& \equiv \sum \limits_{j=1}^l 2^j \ a_{k_j} = \sum \limits_{j=1}^l 2^j \  \tfrac{1}{2} \Re(\mathcal{\tilde N} (e_{k_j}) +1) = \\
& = \tfrac{1}{2} \Re \bigg[ \mathcal{\tilde N} \Big( \sum \limits_{j=1}^l 2^j e_{k_j} \Big)    +  \sum \limits_{j=1}^l 2^j \bigg] =  \\
& = \tfrac{1}{2} \Re \big( \mathcal{\tilde N} (e_{\text{packed}})    +  2^{l+1} -1 \big),
 \end{aligned}
\label{eq:tranf_packed_ak_bk}
\end{equation}
where $e_{\text{packed}} = \sum_{j=1}^l 2^j e_{k_j}$ is a vector composed of a bundle of unit vectors $e_{k_j}$ encoded into the consecutive bit-planes of its integer elements. 
From Eqs.~(\ref{eq:transf_ak_bk}),~(\ref{eq:tranf_packed_ak_bk}) one may easily determine that the
computational complexity of calculating the whole bundle of patterns is identical as in the case of calculating a single pattern. 
The number of patterns, which may be calculated with a single run of the modified noiselet transform, is limited only by the bit-width of the integer variables used in the implementation of the transform.

To illustrate the efficiency of the modified noiselet transform, we note that our C++ implementation of the transform generates around $110$ bundles of patterns of resolution $512 \times 512$ per second or $250$ bundles of patterns of resolution $256\times 256$ per second on a mid-range laptop. For patterns bundled into packages of 24, as is the case in our experiment, this is respectively $2700$ or $6000$ patterns per second.
This speed already includes all the operations on the computer graphical objects and hardware necessary to display the patterns.

\section{Experimental Results}
We will now demonstrate the experimental results of reconstructing an image captured by a SPC. We apply binary sampling equivalent to noiselet sampling according to the procedure proposed in Sections~\ref{sec:patterns} and \ref{sec:patterns_real_time_gen}.

\subsection{Design of the experimental setup}
The schematic of our experimental system matches the one shown in  Fig.~\ref{fig:schemat}. 
We use a DMD light modulator (TI DLP LightCrafter 4500) integrated with an RGB LED light source  and optical lens to project the sampling patterns onto the object plane.
 The DMD consists of $912\times1140$ square micromirrors organised into a diamond grid, i.e. a square grid rotated by $45^\circ$ with respect to the boundaries of the DMD. Out of this array, we use a square sub-area of $512\times512$~pixels oriented along the edges of the pixels, so that the boundaries of the area form straight lines. All the patterns are transformed in order to be displayed in this area, which ensures accurate projection of the patterns into a square grid, without distortions resulting from the diamond pixel layout of the device. A similar approach was previously reported in \cite{Lancis_dual_mode_2016}.
 The single-pixel detector consists of a photodiode  integrated with an on-chip transimpedance amplifier (TI OPT-101P) with peak sensitivity wavelength of $650$~nm. The analog signal is digitized with a $16$-bit A/D converter (NI USB-6003 100kS/s multifunction DAQ) and streamed to a PC via USB port. Data acquisition is controlled with a LabView routine.

The DLP displays binary patterns consisting of arbitrarily selected bit-planes of 24-bit RGB images. These images may be either first stored in the internal flash memory or streamed via HDMI video port. The latter solution is more convenient, since it does not impose memory restrictions and allows for a more flexible choice of patterns, by generating them in real time during the measurements instead of preparing and uploading them beforehand. 
  We have developed a dedicated fast routine in C++ based on modified noiselet transform (see Section~\ref{sec:patterns_real_time_gen}) to generate noiselet-based patterns~(Eq.~(\ref{eq:patterns_mod_noiselets_relation})) 
  bundled together in packages of $23$ (with one bit left for synchronization) into the bit-representations of RGB images. Such images are transferred through HDMI port to the DLP and then displayed in sequence, bit by bit.
Our system is capable of displaying binary patterns at the maximum rate of $1440$~Hz ($24$~bits $\times\ 60$~Hz of the video rate). 
However, the actual speed of displaying the patterns in the experiment is set to $240$~Hz in order to preserve high signal to noise ratio and to ensure stability of the data acquisition.

The accuracy of the measurements is an important issue for the SPC imaging. Since all the noiselet-based binary patterns of the same resolution have the same total brightness, the standard deviation of the measurements taken with different patterns is usually at least two orders of magnitude smaller than their average value.
Therefore, in order to increase the signal to noise ratio and to reduce the influence of both the background signal from the dark pixels and the light intensity variations over time, we use the technique of differential measurements, involving displaying both the patterns and their binary negations. 
In a different SPC design, similar effect could be obtained without doubling the number of measurements, by introducing a second detector to measure synchronously the total intensity of light reflected from the DMD.

Finally, we recover the image from the measured data by solving the BPDN problem (see Eq.~(\ref{eq:min_l1_noise})) with the use of the SPGL1 package \cite{SPGL1-art2011}. 
In the worst case scenario of recovering an image from measurements with 50\% elements of the noiselet basis, the optimisation takes approximately $3$~s for an image with resolution $256\times 256$ (in Matlab, using a PC with a single eight-core processor). In the case of reconstructing the image from the entire noiselet basis, a straightforward approach of calculating the inverse noiselet transform  may be applied, reducing the time of recovery to $0.03$~s.


\begin{figure}[t!]
	\includegraphics[width=4cm, height=3.5cm]{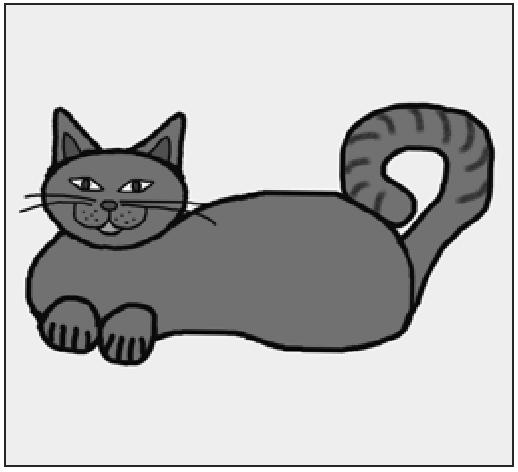}  
	\includegraphics[width=4.8cm, height=3.5cm]{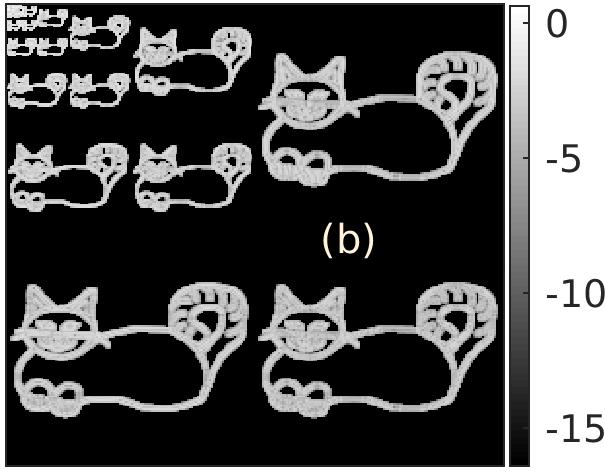}  \\
	\includegraphics[width=4cm, height=3.5cm]{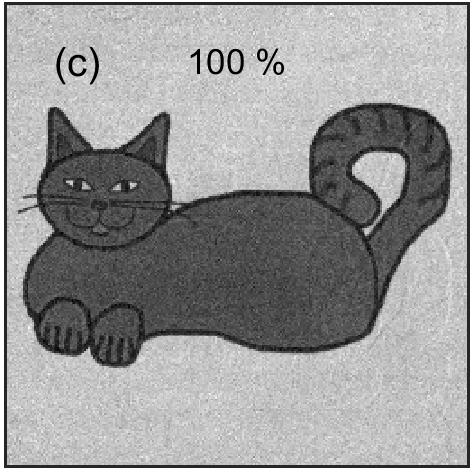}  
	\includegraphics[width=4cm, height=3.5cm]{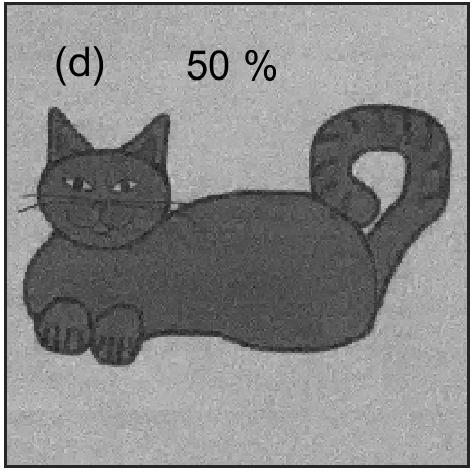}  \\
	\includegraphics[width=4cm, height=3.5cm]{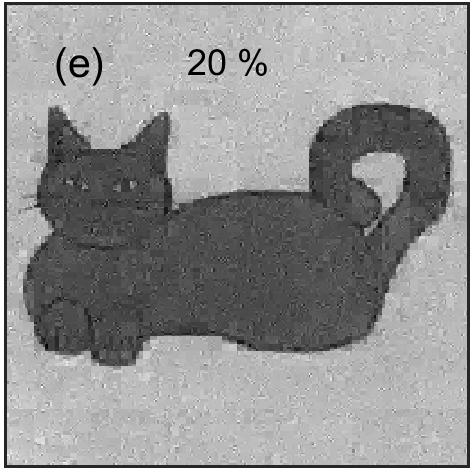}  
	\includegraphics[width=4cm, height=3.5cm]{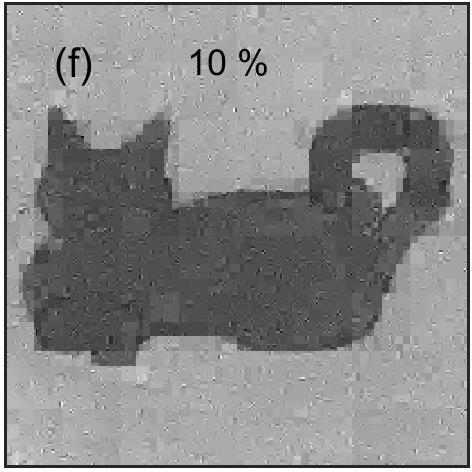}  \\
	\includegraphics[width=8.2cm, height=5cm]{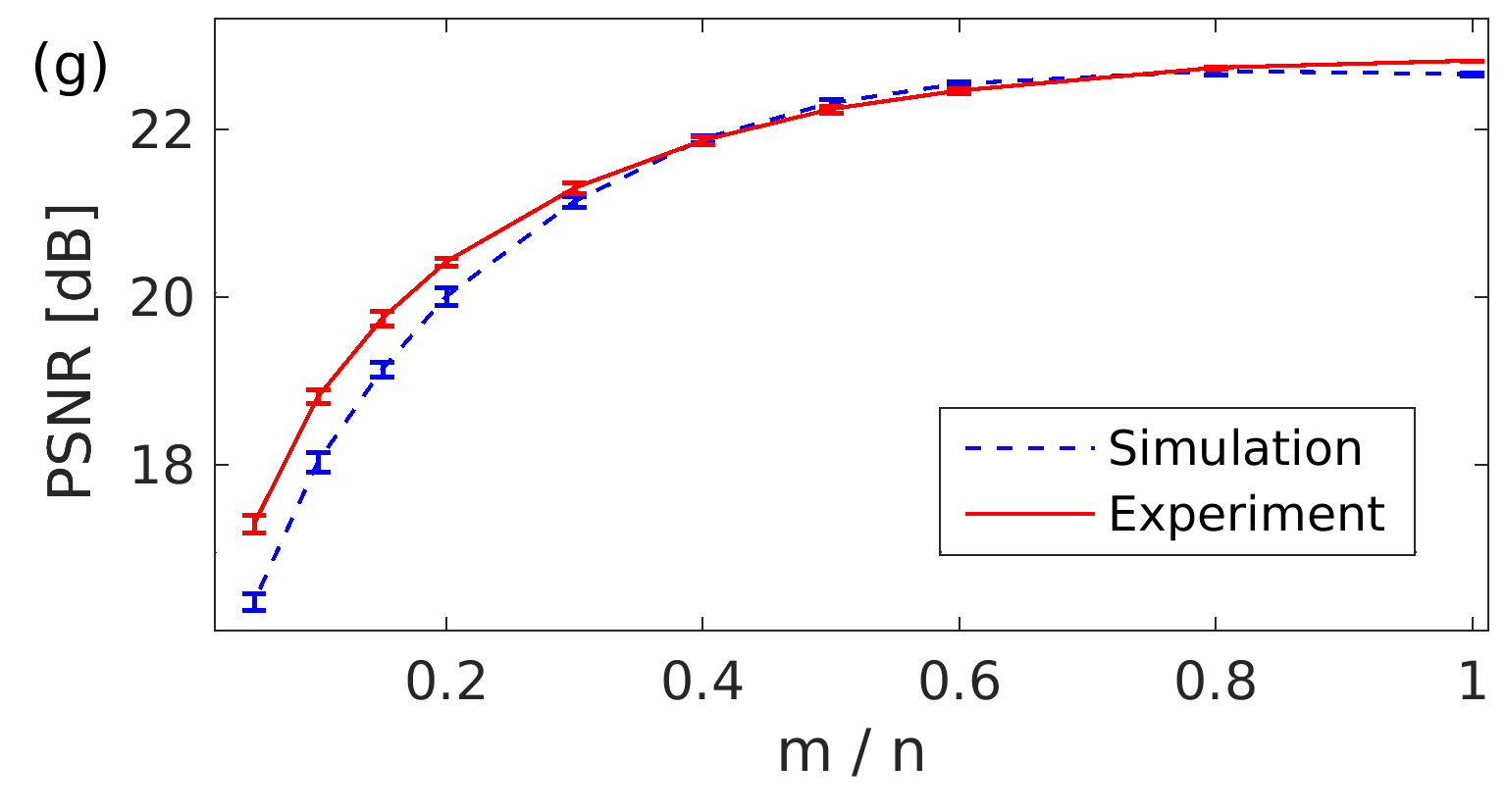}  
\caption{Experimental demonstration of compressive imaging with SPC using noiselet-based sensing patterns with resolution $256 \times 256$: (a)~the original image, (b)~two-dimensional Haar wavelet transform of (a) (logarithmic scale), 
(c-f)~recovery of the image from the experimental data using different sampling ratios $m/n$ between $100\%$~(a) and $10\%$~(d), 
(g)~experimental and theoretical PSNR as a function of the sampling ratio $m/n$. The errorbars in (g) refer to the standard deviation calculated over 40 (for $m/n \leq 0.3$) or 20 (for $0.3<m/n<1$) randomly chosen sets of the sampling patterns with the same value of the sampling ratio $m/n$.
}
\label{fig:Eksperyment}
\end{figure}

\subsection{Results}

The original image used for the experiment is presented in the Fig.~\ref{fig:Eksperyment}(a). 
The image is well compressible in the Haar wavelets, as shown in Fig.~\ref{fig:Eksperyment}(b), with only 24\% of non-zero Haar coefficients. Further lossy compression is also possible, with the peak signal to noise ratio (PSNR, see Eq.~(\ref{eq:reconstruction_error})) of the compressed image on the order of $40$~dB when 8\% of the largest Haar coefficients are preserved.

The image is sampled with the noiselet-based real binary patterns of the resolution $n = 256 \times 256$.
The recovery of the image is repeated using several different values of the sampling ratio $m/n$ from the range between 5\% and 100\%, where 100\% corresponds to the sensing matrix $\Phi$ consisting of all possible noiselets from a single noiselet matrix. 
Several examples of reconstructed images obtained with different values of the sampling ratio are presented in Figs.~\ref{fig:Eksperyment}(c)-(f).
Additionally, in Fig.~\ref{fig:Eksperyment}(g) we present how the sampling ratio influences the accuracy of reconstruction of the image. 
To measure the quality of the reconstruction, we use the PSNR defined as:
\begin{equation}
\textsc{PSNR} =  10 \log_{10} \Bigg( \frac {[\max (X)]^2} {MSE(\tilde X)}  \Bigg),
\label{eq:reconstruction_error}
\end{equation} 
where $\tilde X$ represents distribution of brightness in the reconstructed image, $\max (X)$ stands for the peak brightness of the original image, and $MSE(\tilde X) = \tfrac{1}{n} \sum_{i=1}^n (\tilde X_i - X_i)^2$ is the mean squared error of the reconstructed image as compared to the original one. Before calculating the PSNR, both images have been registered and normalised in order to match their intensity levels.
We note that PSNR is not a deterministic function of the sampling ratio - it depends on the specific patterns used in the measurement or simulation. Therefore, in Fig.~\ref{fig:Eksperyment}(g) we show also the standard deviation of PSNR calculated over a number of randomly chosen sets of patterns with the same value of the sampling ratio.


 

We define the experimental noise as the difference between the measured and calculated values of $\tilde Y$.
The standard deviation of the experimental noise is on the order of $0.0004$ of the mean value of the measured signal and it is by approximately two orders of magnitude lower than the peak-to-peak amplitude of the measurements taken with different patterns. 
By introducing additive Gaussian white noise to the theoretical model of the measurement, we obtain a good agreement between the simulated compressive measurement and the actual experimental results (see Fig.~\ref{fig:Eksperyment}(g)). 
 

\section{Conclusions}
We have proposed theoretically and validated experimentally an efficient method of using complex-valued and non-binary noiselet functions for object sampling in single-pixel cameras with binary spatial light modulators and incoherent illumination. Minimal mutual coherence of discrete noiselets and Haar wavelets makes this pair of bases an essential choice for the sensing and compression matrices in compressed sensing with single-pixel detectors. Indeed, most real-world images are compressible in the Haar basis. The proposed method allows to determine $m$ noiselet coefficients from $m+1$ binary sampling measurements. 
Moreover, we have proposed a modification to the complex fast noiselet transform, which enables computationally-efficient generation of the binary noiselet-based patterns using only operations of summation and subtraction on integer variables. 
Further acceleration is obtained by utilising the bit representation of a $k$-bit integer to calculate a bundle of up to $k{-}2$ patterns without any additional computational cost as compared with generating only a single pattern.
The efficiency of the proposed method allows for generating patterns in real time on a PC or even on a single-board computer.



\section*{Funding Information}

\textbf{Funding.} We acknowledge financial support from the National Science Centre, Poland grant UMO-2014/15/B/ST7/03107 and European Union Seventh
Framework Programme (FP7/2007-2013, grant agreement no 316244).

\bibliographystyle{unsrt}
\bibliography{biblio}


\end{document}